\documentclass[11pt,a4paper]{article} 
\usepackage{caption}         %
\usepackage{charter}         
\usepackage{graphicx}        
\usepackage{amssymb,amsmath} 
\usepackage{amsthm}          
\usepackage{multicol}        
\usepackage{multirow}
\usepackage{url}             
\usepackage{enumitem}        
\usepackage{marvosym}        
\usepackage{wrapfig}         
\usepackage[T1]{fontenc}     
\usepackage{mathptmx}        
\usepackage{datetime}        
\usepackage{fancyhdr}        
\newdateformat{mydate}{\monthname[\THEMONTH] \THEYEAR}          
\usepackage[pdfpagemode=FullScreen, colorlinks=false]{hyperref} 
\usepackage{algorithm}
\usepackage{algorithmic}
\usepackage{booktabs}        
\usepackage{array}
\usepackage{courier}
\usepackage{hyperref}                                      %
\hypersetup{
    colorlinks = true,
    citecolor  = blue,
    linkcolor  = blue,
    urlcolor   = blue
}

\newcommand{\HorRule}[1]{\noindent\rule{\linewidth}{#1}}

\newcommand{\SectionTitle}[1]
{
    \HorRule{1pt}
    \usefont{T1}{ptm}{b}{n}               
    \vspace{3pt}\Large #1\vspace{1pt}    
    \par \normalsize \normalfont
    \HorRule{1pt}
}

\newcommand{\NewsTitle}[1]
{
    \begin{center}
    \usefont{T1}{ptm}{b}{n}               
    \vspace{14pt}\Large #1\vspace{1pt}    
    \par \normalsize \normalfont
    \end{center}
}

\newcommand{\NewsAuthor}[1]
{
    \begin{center}
    \usefont{T1}{ptm}{n}{n}               
    \textsc{#1} \vspace{4pt} 
    \par \normalfont
    \end{center}
}

\graphicspath{{./figs/}}
\usepackage{tabularx}
\usepackage{xcolor,colortbl}
\usepackage{color}
\usepackage{subfig}
\usepackage{cleveref}
\usepackage{filecontents}
\usepackage{pgfplots}
\usepackage{pgfplotstable}
\pgfplotsset{compat=newest}

\usepackage{enumitem}
\usepackage{textcomp}

\begin{document}




\SectionTitle{Technical Articles}
\setcounter{section}{0}
\setcounter{figure}{0}
\graphicspath{{./figs/}{./figs/item-template/}}
\renewcommand\thesubsubsection{\Alph{subsubsection}}
\NewsTitle{Aerial Manipulation Using a Novel Unmanned Aerial Vehicle Cyber-Physical System}
\NewsAuthor{Caiwu Ding$^{1}$, Hongwu Peng$^{2}$, Lu Lu$^{1}$, and Caiwen Ding$^{2}$\\
$^{1}$Department of Mechanical and Industrial Engineering, New Jersey Institute of Technology\\
$^{2}$Department of Computer Science and Engineering, University of Connecticut}

\begin{abstract}

Unmanned Aerial Vehicles(UAVs) are attaining more and more maneuverability and sensory ability as a promising teleoperation platform for intelligent interaction with the environments. This work presents a novel 5-degree-of-freedom (DoF) unmanned aerial vehicle (UAV) cyber-physical system for aerial manipulation. This UAV's body is capable of exerting powerful propulsion force in the longitudinal direction, decoupling the translational dynamics and the rotational dynamics on the longitudinal plane.  A high-level impedance control law is proposed to drive the vehicle for trajectory tracking and interaction with the environments. In addition, a vision-based  real-time target identification and tracking method integrating a YOLO v3 real-time object detector with feature tracking, and morphological operations is proposed to be implemented onboard the vehicle with support of model compression techniques to eliminate latency caused by video wireless transmission and heavy computation burden on traditional teleoperation platforms.

\end{abstract}

\section{Introduction}

\vspace{-.1in}
Unmanned Aerial Vehicles (UAVs) are maturing as a multi-disciplinary technology platform, their inherent aerial capabilities have been further developed and augmented, along with the advancement of powerful and robust perception techniques for aerial vehicles,  UAVs are surpassing the traditional role as passive aerial observation platforms and able to actively interact with the environments to address aerial manipulation tasks in hard-to-reach or dangerous places. The tasks currently solved by aerial manipulators range from grasping~\cite{thomas2014toward}, fetching~\cite{kim2016vision}, writing, peg-in-hole~\cite{park2018odar}, to transporting arbitrary objects. However, all of these tasks are at an entry level and require only very basic tracking and interaction between UAV and the environments. 

Real engineering problems in the daily life often involve sophisticated motion, contact and force control, also real-time precision target identification and tracking for environment perception. But most of these issues still remain hardly achievable under the traditional design, sensing and control of UAV systems, for example: inspection and failure detection of infrastructure like bridges or manufacturing plants, physical interactions through tools like grinding, welding, drilling, for construction projects or maintenance tasks in dangerous or harmful places~\cite{9363592}.

In the aspect of aerial maneuverability, independent forces and torques must be exerted in certain degrees of freedom(DoF) to the environments to achieve a successful aerial manipulation operation, while the rest of DoFs of the end-effector of an aerial teleoperation platform should be accurately position-controlled~\cite{9250626}. However, traditional quadcopters are highly underactuated~\cite{yu2016global,ding20186}, which means the translational motions on the horizontal plane are tightly coupled with rotational motions. Therefore, accurate position-level control and forces/torques control are impossible to be implemented simultaneously. In comparison to the traditional multirotor approaches, we can resolve the underactuation issue by proposing a novel tilting-rotor multirotor with an elegant and concise structure design, which is capable of generating 5-DoF thrust forces and torques, and completely avoiding inefficient force cancellation between rotors. 

For the target identification and tracking of UAVs navigating in a large open space, a target detection-tracking approach has to be followed. Early solutions to object detection tasks depended on traditional machine learning methods, i.e., feature-based manual methods. The difficulty with the traditional approaches is that it is necessary to choose which features are important in each given image. As the number of classes to classify increases, feature extraction becomes more and more cumbersome~\cite{walsh2019deep}. Moreover, most of them are verified in low and medium density images and they usually need to be changed according to the specific situations~\cite{zhang2020applications}, thus not suitable for applications in unstructured environments where illumination variations, partial occlusions, background clutter and shape deformation would occur. A deep learning model is trained based on the given data, and can automatically work out the most descriptive and salient features with respect to each specific class of object~\cite{nweke2018deep}. It has been demonstrated that in many object detection applications, deep learning performs far better than traditional algorithms~\cite{sermanet2013overfeat}. In this work, an onboard vision-based real-time guidance scheme is developed by dividing the identification and tracking problem into three parts: 1)Detect a target with a real-time YOLO v3 object detector,
2)2D transformation tracking using KLT tracker, 3)Extract position of the point of interest with morphological image processing method.

\vspace{-.2in}
\section{System Architecture And Mechanical Design}\label{sec:control}
\vspace{-.1in}

\begin{figure}[t]
\begin{center}
\centerline{\includegraphics[width=0.6\columnwidth]{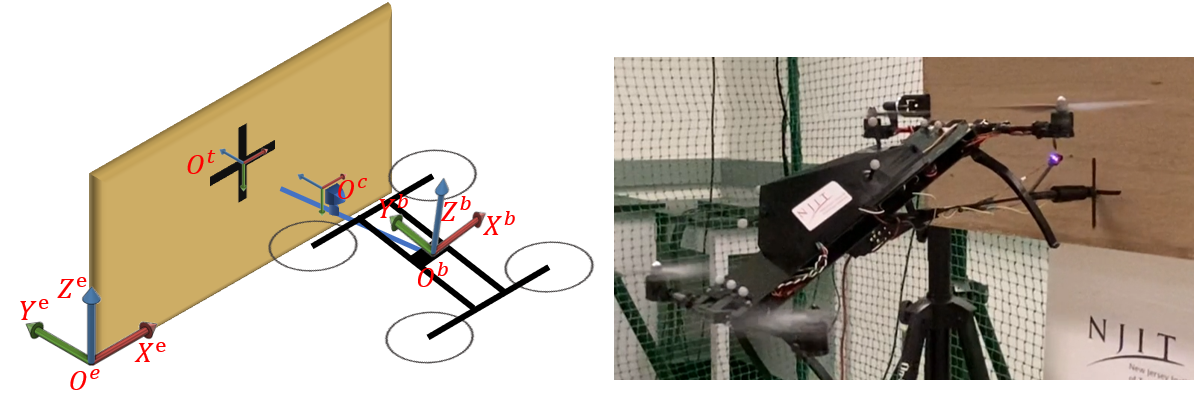}}
\caption{Prototype of the novel UAV cyber-physical system built in the Assistive and Intelligent Robotics Laboratory at NJIT: image stream from eye-in-hand camera is fed into the vision-based target identification and tracking module; target position and orientation are sent to the impedance control law for motion/force control.}
\label{structure}
\end{center}
\end{figure}

This section is to demonstrate the architecture and mechanical design of the developed tilting-rotor UAV, furthermore, to clarify the ability of independent forces/torques control, and independent position control in a mechanical design view. In addition, advantages of the UAV design are elaborated compared with previous aerial teleoperation platforms.

To perform aerial manipulation with traditional coplanar multirotors like quadcopter, hexacopter, and octocopter, aerial robots are equipped with a n-DoF robtic arm~\cite{6869148}. However, this solution comes with severe drawbacks.  Firstly, a robotic arm strongly decreases the payload and flight time of an aerial vehicle due to its weight. Secondly, the system is much more complex mechanically, thus, it requires more maintenance and repairing efforts. Besides, lateral/longitudinal forces in body frame, which cannot be provided by the aerial platform itself, have to be generated through the dynamical/inertial coupling between the arm and the aerial robot, which is extremely hard to achieve in real-world conditions.

To solve both the underactuation problem and issues associated with robotic arms at once, many researchers have developed multirotor UAVs with fully-actuated aerial mobility. These designs use more than 6 rotors facing different directions~\cite{park2018odar}, or include one/two extra servomotor for each rotor~\cite{ding2018energy}, to achieve 6-DoF actuation. However, the inclusion of a large number of motors or servo parts makes the designs mechanically more complicated, heavier, and the inefficient internal force cancellation can hardly be avoided~\cite{ding2018modeling}.

The novel tilting-rotor multirotor UAV introduced in this work employees only two additional servomotors compared to traditional quadcopters, to actuate the two pairs of rotors on the vehicle. Specifically, the two pairs of rotors are mounted on two independently-actuated arms placed at both sides of the vehicle in an ``H'' configuration, as shown in Figure \ref{structure}. Each arm is driven by a single servomotor, and carries two rotors on its two ends. Hence, the 5-DoF forces and torques can be generated by tilting the two arms.

\vspace{-.2in}
\section{Motion/Force Control And Real-time Target Tracking}\label{sec:control}
\vspace{-.1in}

\begin{figure}[t]
\begin{center}
\centerline{\includegraphics[width=0.6\columnwidth]{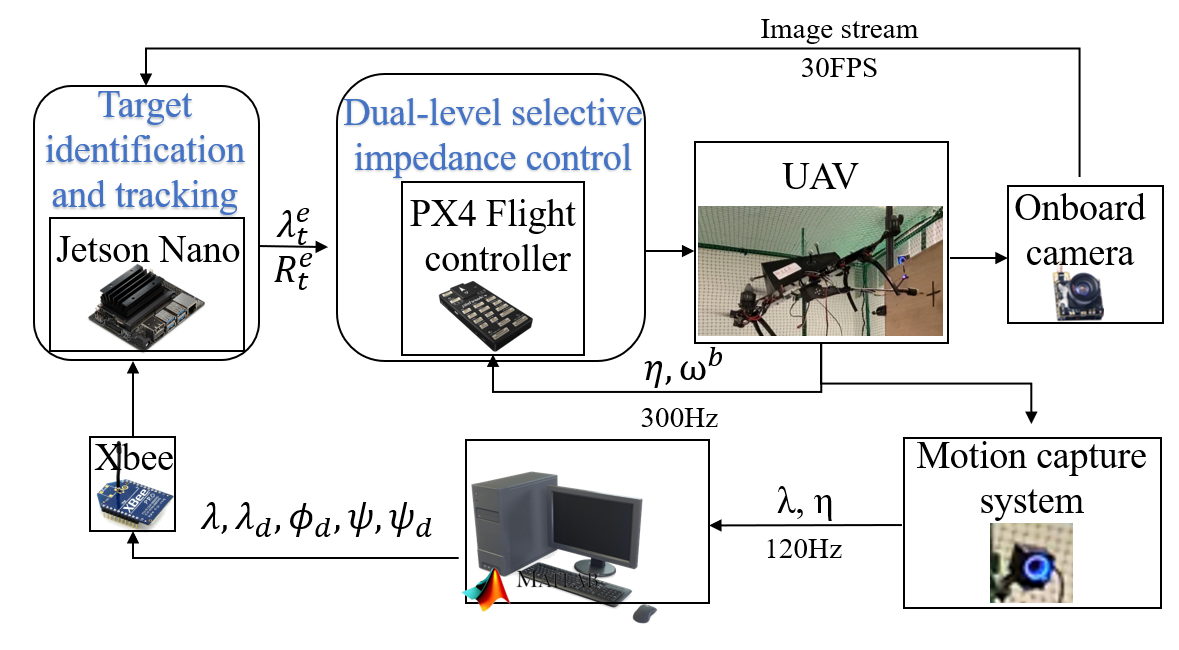}}
\caption{An overview of the UAV aerial manipulation system: the target identification and tracking algorithm is running onboard on the Jetson nano micro-controller to capture the target position and orientation in real-time; the dual-level impedance control algorithm is running on the PX4 flight controller at 300$Hz$.}
\label{overview}
\end{center}
\end{figure}

This section presents the motion/force controller for position tracking and contact force regulation and the vision-based target identification and tracking system, these two modules are integrated together on the proposed UAV cyber-physical system for performing precision aerial manipulation tasks in unstructured environments, as shown in Figure \ref{overview}. 

\subsection{Motion/Force Control.}

A dual-level cascaded control architecture is deployed for flight control. The low-level attitude controller uses a PID controller to generate updates of the required body-frame torques $\tau^{b}$ to accurately regulate the attitude angles, while a selective impedance controller is implemented at high level to indirectly regulate the contact force with environments $F_p^b$ and track the target position captured by the vision-based target identification and tracking module. Finally, the five body-frame force and torque inputs  will be used to generate the propeller thrust force commands and the tilt angle values to be fed into the rotors and servos respectively. In the fact that the PID control method has been widely applied to various commercial UAVs for the low-level attitude regulation, and has been developed as a standard approach with guaranteed stability and robustness verified in real-world applications, this work will focus on the introduction of high-level impedance controller for motion/force control of the proposed novel UAV.

Firstly, the total thrust force generated by the vehicle can be expressed in body frame as $F_p^b=[0 \ \ F_{py}^b \ \ F_{pz}^b]^T$. Defining $F_1$, $F_2$, $F_3$, $F_4$ as the absolute values of the thrust forces of the four propellers and $\alpha$, $\beta$ as the front and rear tilting angles, the $F_{py}^b$ and $F_{pz}^b$ can be expressed as:
\begin{equation}\label{F_body}
    \begin{array}{rcl}
    F_{py}^b & = & (F_1  + F_4)S_\alpha + (F_2 + F_3)S_\beta,\\
    F_{pz}^b & = & (F_1  + F_4)C_\alpha + (F_2 + F_3)C_\beta.
    \end{array}
\end{equation}

In the same way, the total torque is expressed as $\tau^{b}
=\begin{bmatrix}\tau_x^b&\tau_y^b&\tau_z^b\end{bmatrix}^T$.
Defining $l_1$ as the half distance between the two rotors on each tilting axis, $l_2$ as the half distance between the two tilting axes, the components of $\tau_p^b$ can be expressed in body frame as
\begin{equation}\label{tau_body}
    \begin{array}{rcl}
    \tau^b_{x} & = & (F_1 + F_4)l_2C_\alpha - (F_2 + F_3)l_2C_\beta, \\
    \tau^b_{y} & = & (F_1  - F_4)(C_\alpha l_1+S_\alpha k) + (F_2 - F_3)(C_\beta l_1-S_\beta k), \\
    \tau^b_{z} & = & (F_4 - F_1)(S_\alpha l_1 -C_\alpha k) + (F_3 - F_2)(S_\beta l_1 + C_\beta k).
    \end{array}
\end{equation}

It's defined that $\lambda^t=[x^t\ y^t\ z^t]^T$ is the coordinate vector of vehicle's position $O_b$ expressed in the target frame whose origin coincides with the  target point and $Z^t$ axis is perpendicular to the workpiece surface. 

According to the Newton–Euler equations, differentiating $\lambda^t$ twice, yields the following translation dynamic model
\begin{equation}
    M\ddot{\lambda}^t=F^t_p+F^t_g+F^t_c,
\end{equation}
where $F^t_p$, $F^t_g$, $F^t_c$ are the total thrust force, gravity force, and contact force represented in target frame, respectively. 

The selective impedance controller is designed as
\begin{equation}\label{impedance_control}
    F^t_p=-F^t_g-F^t_c+M\ddot{\lambda}^t_d -C\dot{e}^t_{\lambda} - Ke^t_{\lambda},
\end{equation}
where $e^t_{\lambda}=\lambda^t-\lambda^t_d$ is the position tracking error in target frame, $\lambda^t_d$ is the desired position, $C=diag(C_x,C_y,C_z)$ and $K=diag(K_x,K_y,K_z)$ are diagonal matrices consisting of all the virtual damping and spring parameters along $X^t$, $Y^t$, $Z^t$. With the proposed impedance controller design,  the following impedance model relating the position tracking error to the contact force can be derived:
\begin{equation}
    M\ddot{e}^t_{\lambda}+C\dot{e}^t_{\lambda}+Ke^t_{\lambda}=F^t_c.
\end{equation}

The damping and spring parameters for different axes can be tuned according to the desired hybrid motion/force control objective.

After obtaining the target frame thrust force $F^t_p$, the body frame thrust force can be calculated as $F^b_p=R^b_tF^t_p$, then $F^b_{py}$, $F^b_{pz}$ can be derived directly. Finally, by setting $\alpha=\beta$ and solving the five equations in (\ref{F_body}) and (\ref{tau_body}), the four individual rotor thrust forces and the tilt angle can be obtained.

\subsection{Real-time Target Tracking.}



\subsubsection{Deep Neutral Network (DNN) Based Objection Detection.}
The most important task for the UAV target tracking mission is real-time object detection. The DNN-based object detection models are first trained on labeled images and are then used to predict the targets' bounding box and classification. The state-of-the-art DNN based object detection models can be divided into two classes: region proposal network (RPN) based detection method and single-stage detection method.

\textbf{RPN based detection method.} The RPN-based detection method requires one stage to extract the region of interest (RoI), and another stage to predict the classification result and bounding box. The famous RPN based methods include SPP network~\cite{he2015spatial}, R-CNN \cite{girshick2014rich} and its' variant Faster R-CNN \cite{ren2015faster}. Although the R-CNN gains significant improvement in its' performance in recent years, its' inference speed is still limited by the high computation burden of the region proposal stage. Thus, R-CNN not suitable for the UAV's real-time detection application. 

\textbf{Single stage detection method.} The single-stage detection eliminates the need for the region proposal stage and proposes an end-to-end structure to predict bounding boxes and class probabilities by a single evaluation step. The typical single-stage detection networks include the You Only Look Once (YOLO) \cite{redmon2016you} network and it's variants YOLOv2 \cite{redmon2017yolo9000}, YOLOv3 \cite{redmon2018yolov3}, and YOLOv4 \cite{bochkovskiy2020yolov4}. The variants of YOLO have improvement on both parameter size and prediction accuracy. Single-Short MultiBox Detector (SSD) \cite{liu2016ssd} is another type of single-stage detection method. It features its' high prediction speed, but it has lower accuracy and is limited to a fixed number of bounding boxes. The challenge of single-stage detection lies in improving the inference speed while retaining good accuracy. For our UAV real-time objection detection application, YOLO and its' variants are more suitable than other DNNs. 

\subsubsection{Inference Acceleration for Real-time Detection.}

In order to achieve accelerating the DNN inference speed, model compression techniques are adopted to reduce the DNN model's parameter size. The major model compression techniques include weight pruning \cite{li2020ftrans,peng2021accelerating, yuan2021improving, peng2022length, qi2021accommodating, huang2021hmc, qi2021accelerating, huang2022automatic, peng2022towards} and weight quantization \cite{ding2017circnn, peng2021binary, peng2021optimizing}. There are unstructured pruning methods and structured pruning methods for the weight pruning technique. The unstructured pruning technique \cite{geng2019o3bnn} prunes out the weight matrix in an irregular pattern and can achieve a higher compression ratio without much accuracy loss. However, due to its irregular memory access pattern, the unstructured pruning can hardly be accelerated on most of the hardware platforms. The structured pruning technique \cite{cai2020yolobile, li2020efficient, yuan2019ultra} constrains the weight matrix to be pruned in a structured and hardware-friendly pattern. For example, block-circulant matrices \cite{ding2017circnn, lu2017evaluating, liao2017energy} can be used for weight representation after pruning. The structured pruning-based hardware implementation achieves better performance due to the higher parallelism achievable by regular memory access patterns and reduced computation burden. 

For most of the objection detection tasks, the convolution layer is the central part of the networks. Fast Winograd algorithm and fast Fourier transform (FFT) algorithm can be used to accelerate the convolution operations and can also accelerate the CNN models with small filters \cite{lavin2016fast}. For example, $3 \times 3$ convolution layers make up $83\%$ of the YOLOv4's weights and $81\%$ of the YOLOv4's computations. Those algorithms bring the computation complexity of convolution operation down from $O(n^2)$ to $O(n \, log \, n)$.  

In order to reduce the communication latency and data transferring latency for UAV applications, object detection tasks need to be deployed on the onboard acceleration platform. Researchers are using the NVIDIA Jetson TX series embedded GPU platform to accelerate YOLO  architecture for real-time object detection \cite{bhandary2017robust, rudiawan2017deep, smolyanskiy2017toward}, and a 20 FPS detection speed is achieved in \cite{smolyanskiy2017toward}. FPGA platform also gains popularity on edge computing application, \cite{ding2019req} implemented YOLO network on Xilinx ADM-7V3 FPGA and achieved a 314.2 FPS detection speed. Simultaneously, the energy efficiency is seven times higher than that of the Jetson TX2 platform. 


\vspace{-.1in}

\section{Conclusion}\label{sec:frame}
\vspace{-.1in}

While UAVs have demonstrated great potential to be employed as teleoperation paltforms for aerial manipulation, the underactuation nature of traditional multirotors, and the latency caused by data transferring and heavy computation burden on traditional aerial teleoperation platforms still hinder the application of UAVs in aerial manipulation. To overcome these problems, we proposed a novel tilting-rotor UAV to cope with the underactuation issue, and an onboard real-time target identification and tracking approach to eliminate latency in the system. Based on the 5-DoF tilting-rotor UAV, a selective impedance control law is designed for motion/force control, which enables the UAV to track the target object whose position and orientation are captured by the onboard vision-based identification and tracking module, and to regulate contact forces between the end-effector and the environments.


{\vspace{\baselineskip}
\bibliographystyle{IEEEtran}
\bibliography{bibligraphy}
}

\clearpage


\end{document}